\UseRawInputEncoding
\documentclass[twocolumn,envcountsect]{svjour3}          
\smartqed  
\usepackage{graphicx}
\usepackage{times}
\usepackage{epsfig}
\usepackage{amsmath}
\usepackage{amssymb}
\usepackage{booktabs}
\usepackage{multirow}
\usepackage[flushleft]{threeparttable}
\usepackage{authblk}
\usepackage[colorlinks]{hyperref}
\usepackage[nameinlink]{cleveref}
\usepackage{mathtools}
\usepackage[misc]{ifsym}

\usepackage{algorithm,algcompatible,amssymb,amsmath}
\renewcommand{\COMMENT}[2][.5\linewidth]{%
  \leavevmode\hfill\makebox[#1][l]{//~#2}}
\algnewcommand\algorithmicto{\textbf{to}}
\algnewcommand\RETURN{\State \textbf{return} }
\makeatletter
\newcommand{\distas}[1]{\mathbin{\overset{#1}{\kern\z@\sim}}}%

\newcommand{\otoprule}{\midrule[\heavyrulewidth]}

\DeclareMathOperator*{\argmin}{arg\,min}
\usepackage[printwatermark]{xwatermark}
\usepackage{xcolor}

\newcommand{\distras}[1]{%
	\savebox{\mybox}{\hbox{\kern3pt$\scriptstyle#1$\kern3pt}}%
	\savebox{\mysim}{\hbox{$\sim$}}%
	\mathbin{\overset{#1}{\kern\z@\resizebox{\wd\mybox}{\ht\mysim}{$\sim$}}}%
}
 \usepackage{mathptmx}      
 \journalname{Preprint}
\graphicspath{{./img/}}
\begin{document}
\sloppy 

\title{Non-Euclidean Analysis of Joint Variations in Multi-Object Shapes
}

\titlerunning{Non-Euclidean Analysis of Joint Variations in Multi-Object Shapes}        

\author{Zhiyuan Liu$^{\textrm{\Letter},1}$ \and J\"orn Schulz$ ^2 $\and Mohsen Taheri$ ^2 $ \and Martin Styner$ ^1 $ \and James Damon$ ^3 $ \and Stephen Pizer$ ^1 $ \and J. S. Marron$ ^4 $ 
}
\authorrunning{Z. Liu et al.} 

\date{Received: date / Accepted: date}
		\institute{ Zhiyuan Liu \at
	\Letter \hspace{0.1ex} \href{mailto:zhiy@cs.unc.edu}{zhiy@cs.unc.edu} 
	\and
	J\"orn Schulz \at
	jorn.schulz@uis.no
	\and
	Mohsen Taheri \at
	mohsen.taherishalmani@uis.no
	\and
	Martin A. Styner \at
	styner@cs.unc.edu
	\and
	James N. Damon \at
	jndamon@email.unc.edu
	\and
	Stephen M. Pizer\at
	pizer@cs.unc.edu
	\and 
	J. S. Marron\at
	marron@unc.edu
\and
$ ^1 $ Department of Computer Science, University of North Carolina at Chapel Hill (UNC),  USA\\
$ ^2 $  Department of Mathematics \& Physics, University of Stavanger (UiS), Norway\\
$ ^3 $ Department of Mathematics, UNC,  USA\\
$ ^4 $  Department of Statistics, UNC, USA }

\maketitle

\begin{abstract}
This paper considers joint analysis of multiple functionally related structures in classification tasks. In particular, our method developed is driven by how functionally correlated brain structures  vary together between autism and control groups. To do so, we devised a method based on a novel combination of (1) non-Euclidean statistics that can faithfully represent non-Euclidean data in Euclidean spaces and (2) a non-parametric integrative analysis method that can decompose multi-block Euclidean data into joint, individual, and residual structures. We find that the resulting joint structure is effective, robust, and interpretable in recognizing the underlying patterns of the joint variation of multi-block non-Euclidean data. We verified the method in classifying the structural shape data collected from cases that developed and did not develop into Autistic Spectrum Disorder (ASD).  
\end{abstract}

\section{Introduction}
\label{intro}
Medical objects like human organs are often functionally and spatially interrelated, e.g., brain structures like the hippocampi and caudate nuclei. Utilizing multi-object data can reveal important information that is ignored in a single object analysis. The additional information is expected to lead to a more robust and sensitive statistical analysis \cite{Cerrolaza2019}. In this work, we focus on multi-object shapes and, therefore, capture the geometric interrelation in a multi-object analysis.

In medical applications, the morphological information of brain structures has been studied in relation to different diseases like Autistic Spectrum Disorder (ASD). Several studies have exposed an association between morphological changes of single brain structures and the development of ASD. For example, \cite{murphy2012anatomy,eilam2016neuroanatomical,Katuwal2015} adopted voxel representations of single structures and showed volume and area differences between the structures of ASD and those of the non-ASD group. Nicolson et al. \cite{nicolson2006detection} led a trend towards revealing more subtle correlations by looking beyond volume and area into local shape of the hippocampus.

This trend continues with a large variety of studies analyzing single structures with sophisticated shape models like boundary point distribution models (e.g., \cite{cates2007shape,oguz2008cortical,achterberg2014hippocampal}) and skeletal models (e.g., \cite{jp_diss,schulz2016}). However, from the physiological and statistical perspectives, more gains are expected to follow by jointly analyzing multiple structures. Given $n$ multi-object configurations where each configuration consists, for example, of a hippocampus and a caudate, the joint analysis can decompose the variation in the data into joint variation and individual variation.


A straightforward joint analysis of multi-object shapes is to focus on feature spaces (e.g., (pre-)shape space \cite{bhattacharya2009statistics,dryden2016statistical,srivastava2016functional}). Yet, this approach is challenged by High Dimensional Low Sample Size (HDLSS) problems \cite{liu2017deep,shen2016statistics}. Instead of focusing on feature spaces, our method extracts the joint shape variation from a low dimensional score space, called the \textit{joint variation subspace}. In particular, we map multi-object shapes into that subspace by extending a method called Angle-based Joint and Individual Variation Explained (AJIVE) (\cite{feng2018}), that is designed for Euclidean data, to a  method for multi-object shape data, as discussed in \cref{sec_jive}. We call this proposed method NEUJIVE (short for Non-EUclidean Joint and Individual Variation Explained). Comparing to the previous methods for integrative analysis (e.g., \cite{schwarz2010multiobject,SHEN2014310,shu2020d,khan2019approximate}), our method yields a more sensitive approach that characterizes shape distributions because we have considered the manifold structure of the data. Further, since we extract the joint shape variation space from the score space, the proposed method can alleviate the problems of HDLSS. We also propose to use landmarks derived from skeletal representations, i.e., s-reps \cite{zhiyuan2020}, to represent single objects in order to have both good correspondences on objects' boundaries across configurations and for convenience of alignment. 

We utilize the joint shape variation in classification of ASD and non-ASD groups. Because multiple brain structures relate to ASD in terms of morphological features, as shown in previous research (e.g., \cite{fu2021novel,duan2020subcortical}), we hypothesize that the joint shape variation is driven by the development of ASD. Therefore, the joint variation pattern in the score space that can be interpreted as the configuration-level variation pattern due to the development of ASD becomes informative if our hypothesis is true.

The rest of this paper is organized as follows. In \cref{sec_back} we introduce the formal problem and the related methods. In \cref{sec_methods} we detail our method from three perspectives: (1) how the method works for multi-block Euclidean data; (2) how we extend the method to the non-Euclidean domain and (3) the landmark shape model we developed based on skeletal representations. In \cref{sec_ev} we demonstrate that joint analysis of multi-block non-Euclidean data with our proposed method can extract useful patterns of variation of simulated data. Our method is also verified in applications (including hypothesis testing and classification) to ASD data. The paper concludes with remarks and discussions in \cref{sec_conclude}.



\section{Background}
\label{sec_back}
This section first gives the formal problem. Then we review basic concepts and methods related to the problem.

\subsection{Problem Statement}

To classify the ASD and the non-ASD group, we set up a binary classification model as
\begin{equation}\label{eq_classification}
\hat{y} = \sigma(w^T\hat{X})  
\end{equation} where $ w\in\mathbb{R}^{d\times 1}$ is a learnable weight vector and $T$ means transpose. The design matrix $ \hat{X} \in \mathbb{R}^{d\times n} $ represents the data of $ n $ configurations\footnote{A configuration is a sample represented by a $d$-tuple.} in a certain feature space; $ d $ is the dimension of the feature space. We highlight that in this paper the rows of $\hat{X}$ are features, while the columns are configurations. The output $\hat{y} \in\{0, 1\}^n$ denotes the vector of predicted labels for the $n$ configurations. The symbol $\sigma(\cdot)$ denotes a mapping from $n$ scores to their predicted binary class labels.

As stated above, we hypothesize that the integrated features $\hat{X}$ representing multiple subcortical shapes can facilitate better classification performance between ASD and non-ASD groups. In particular, we want to integrate shape information from the hippocampus and the caudate nucleus because it has been found that the morphology of these two structures correlates with the development of ASD \cite{macduffie2020sleep,richards2020increased,qiu2010basal}. 

In general, each configuration is composed of $ K $ disjoint objects ($K=2$ in this paper). Let $ d_k $ denote the dimension of shape features of the $ k^{th} $ object, where $ k=1, \cdots, K $. The $ k^{th} $ object is thus represented by a matrix block $ X_k$ of dimension ${d_k\times n} $, where $n$ is the number of configurations. By stacking the $K$ blocks on top of each other, we obtain a composite matrix ${X}= [X_1^T, \cdots, X_K^T]^T$ of dimension $(d_1+\cdots+d_K)\times n$. We aim for the low rank component $\hat{X}$ of ${X}$ that represents the joint variation of $X_1, \cdots, X_K$.

The blocks $X_1, \cdots, X_K$ can be obtained in various ways such as from the pre-shape, shape spaces or the score space. These spaces have different mathematical structures. Considering the properties of these spaces, we choose to construct the joint variation component of ${X}$ in the score space that can have Euclidean structures.

\subsection{Spaces for Shape Analysis}
\label{sec_shape_space}
A representation of shape features induces a space in which the data live. For example, landmark-based representations  are often associated with Kendall's shape space \cite{kendall1984}. A shape space is expected to be invariant under translation, rotation and scaling of objects. 

One way to construct such a desirable space for shape analysis is to remove the location, orientation and size information of objects. Particularly, we can align shapes so as to minimize the Procrustes distance between an object and a reference shape (e.g., Procrustes mean shape of unit size \cite{dryden2016statistical}). Assuming a unique solution is achieved, we can transform each shape to the closest shape to the reference shape in the following senses: (a) every transformed shape is centered to remove location; (b) every transformed shape is rotated to minimize the Procrustes distance to the reference shape and (c) each transformed shape is of the same size with the reference shape (e.g., of unit size). It follows from (a) and (c) that the feature vector $x_i$ of a shape is constrained by $||x_i-\bar{x}||^2=1$, where $\bar{x}$ is the origin of the coordinate system. This induced spherical space is also referred to as the \textit{pre-shape space}. In the pre-shape space, each data point lies on an orbit (an equivalent class) of the same shape but different orientations. Minimizing the rotational distance, as described by (b), can be understood as moving each shape along its orbit such that the distance between any two shapes represents the distance between the two orbits. In fact, the orbit-to-orbit distance can also represent the distance in the \textit{shape space} \cite{dryden2019principal}, which is a quotient space with various orientations of a shape identified.

Both pre-shape and shape space interpret data from the perspective of features. Despite the different topology between the pre-shape and shape spaces, these two spaces share the same dual space in which scores of samples are considered. Such a space is called a score space. For a population of $n$ observations, each data point in the score space is a vector in $\mathbb{R}^n$.

\begin{figure}[h!]
    \centering
    \includegraphics[width=\linewidth]{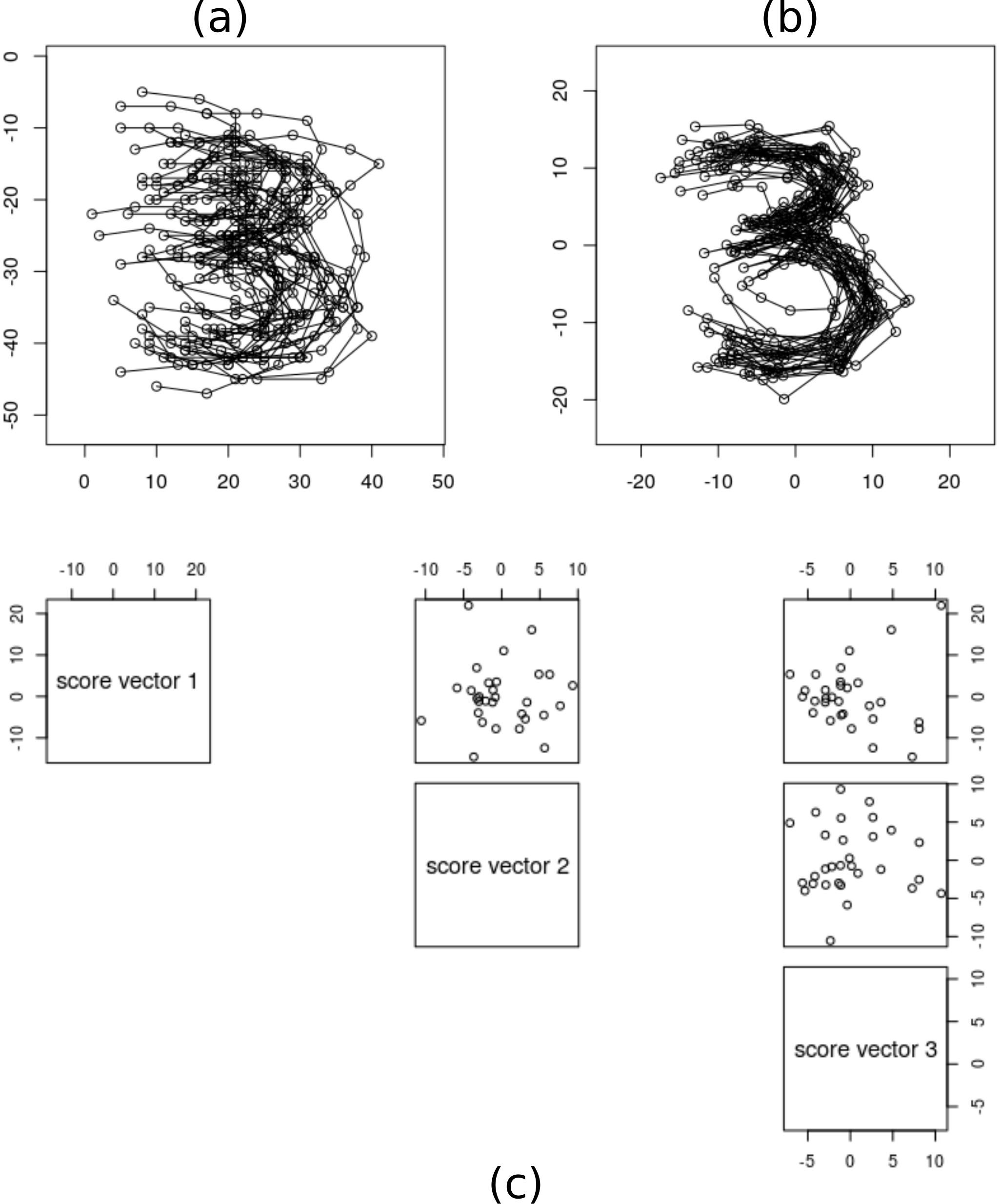}
    \caption{{Different perspectives of a population (30 samples) of handwritten digit ``3'' from \cite{dryden2016statistical}}. (a) The landmarks of all the observations in 2D  space. (b) Aligned shapes from GPA in the 2D feature space. (c) See 30 observations from a score subspace spanned by three score vectors. The three score vectors are, score vector 1 (the vertical axes of the top row), score vector 2 (the horizontal axis of the middle column and the vertical axis of the middle row) and score vector 3 (the horizontal axes of the right column). }
    \label{fig_preshape_shape_space}
\end{figure}

\begin{example}
    \Cref{fig_preshape_shape_space} provides various viewpoints of 30 observations of the handwritten digit ``3''  \cite{dryden2016statistical} from the raw 2D feature space (see \cref{fig_preshape_shape_space} (a)), the aligned configurations (see \cref{fig_preshape_shape_space} (b)) and the low dimensional score subspace (see \cref{fig_preshape_shape_space} (c)). Each observation is represented by 13 landmarks in 2 dimensions. The alignment with Generalized Procrustes Analysis (GPA) removes the position, size and rotation of every shape in pre-shape space, as shown in \cref{fig_preshape_shape_space} (b). From the principal components of the aligned shapes, we obtain the scores of the observations and present them in a low dimensional space, as shown in \cref{fig_preshape_shape_space} (c).
\end{example}

The feature space and scores provide different perspectives in understanding the data. From the feature space, we can obtain knowledge about how subjects distribute with respect to the measured features. The data points in the score space, however, have indirect connections with the measured features. Rather, the score space tells how the data objects vary due to latent factors. 

Multiple blocks in ${X}$ can share a score subspace that is referred to as a \textit{joint variation space} $J$. The basis vectors that span $J$ tend to reflect the common factors that drive the joint variation, neglecting the heterogeneity of features (e.g., units, dimensions, scales) across the blocks. Take the multi-object shape analysis as an example, the joint variation space of neighboring objects is invariant to linear transformations of the features. Rather, the structure of the joint variation space depends on the degree of correlation between objects.

\paragraph{Related work.} A straightforward approach to obtain the basis for the joint variation space is via a multi-block Principal Component Analysis (PCA) \cite{Wold2005} on the composite matrix ${X}$. However, the principal directions resulting from this approach can be dominated by the block of relatively larger variability, dismissing the joint variability. 

Alternatively, Lock et al. \cite{lock2013joint} proposed Joint and Individual Variation Explained (JIVE) to decompose data into three structures, namely, the joint, individual, and residual structures. With this decomposition model, the method iteratively decomposes the data via minimizing the residual components. More recently, Feng et al. \cite{feng2018} proposed a variant of JIVE to address the computational burden. This work inspires us to look for principal vectors in scores spaces ($ \mathbb{R}^n $) rather than canonical loadings in feature spaces (see the following discussion). However, this method has no consideration for the manifold structure of multi-object configurations. 

Compared to the above methods, Canonical Correlation Analysis (CCA) \cite{HOTELLING1936} maximizes the correlation over all possible directions, yielding pairs of canonical loadings $ \{u_k\in U_k|k=1,\cdots, K\} $, where $ U_k $ is the left singular matrix of $ X_k $. However, the canonical loadings are not well-defined in HDLSS problems. Likewise, Partial Least Squares (PLS) \cite{Wold2004} directions are also found in feature spaces. It differs from CCA in that the objective function is to maximize covariance instead of correlation. However, the resulting scores can be affected by structured variation in an individual set of measurements \cite{Trygg2003}.

To sum up, in this section we have explained the difference between the feature space (i.e., pre-shape and shape space) and the score space of a data block. Moreover, we have reviewed some methods of multi-block data analysis. Although these methods approach the joint variation from the various  spaces as above mentioned, none of them has considered the manifold structure of shape data.

\subsection{Non-Euclidean Statistics}
\label{sec_non_euc}
As above reviewed, many existing joint analysis methods are built upon the Euclidean metric. Spaces for shape analysis, however, are often curved spaces. It is the curvature of the space that changes the metric and that affects the measurement of distances and similarities between shapes \cite{huckemann2010intrinsic}. Therefore, the joint analysis of shapes should be sensitive to the metric in order to produce satisfactory results.

A well-known solution to address the non-Euclidean metric is to \textit{Euclideanize} the data \cite{steve2019}. Since a population of biological shapes often distributes within a singularity-free curved subspace of rather low curvature \cite{klingenberg2020walking}, a tangent space can well approximate the manifold structure for the population. Therefore, curves and points on the manifold can be approximated by curves (or lines) and points in the tangent space. For example, great circles (also geodesics) through the point of tangency on the manifold are mapped to straight lines in the tangent space, while small circles (also non-geodesics) on the manifold can be mapped to circles in the tangent space. Hence, the mean and variation of the population can be well approximated in the tangent space of the manifold, on which the data live.

The fitted tangent space (e.g., at the Fr\'echet mean) is a Euclidean vector space in which linear algebra is applicable. Some research \cite{fletcher2004principal} fits  linear models to the variation patterns of data in the tangent space, resulting in distributions along great circles on the manifold. However, Jung et al. \cite{sungkyu2012} have found that not every population distributes along great circles. Moreover, they have proposed the method Principal Nested Spheres (PNS) to decompose the data into nested subspheres. The lowest dimension of subspheres that is non-decomposable (i.e., $\mathbb{S}^0$) gives a notion of the mean value of the population. 

Both the fitted tangent space and the nested subspheres yield Euclidean representations of a population. The nested subspheres method has more theoretical merits in that it provides a universal model for different (including both great and small circles) modes of variation on the manifold. Moreover, the nested subspheres method achieves a backward mean that is more representative of the population.

\begin{example}
	To illustrate the difference between various methods of non-Euclidean statistics, let us consider a toy example in which $ n=50 $ samples are distributed on a unit sphere $ \mathbb{S}^2 $ as shown in \cref{fig_toy_non_euclidean}. The data points are shown as the dark grey dots on $ \mathbb{S}^2 $ in \cref{fig_toy_non_euclidean}. The data are generated along a circle on the tangent space at the north pole. After projecting the data onto the sphere, we rotate the data away from the north pole. Thus, the data generation can be written as 
	\begin{equation}\label{toy_exp_non_euclidean}
X(\theta) = g(\varphi^{-1}(e^{i\theta} + \epsilon))
\end{equation}
where $\theta$ is a uniformly distributed variable $ \theta\distas{} \text{Uniform}(0, \frac{3}{2}\pi)$. The additive term $ \epsilon\distas{i.i.d} N(0, \sigma^2)$ is random noise. The operator $ \varphi^{-1} $ is an exponential mapping centered at the north pole. The operator $ g $ rotates data on the sphere.

\Cref{fig_toy_non_euclidean} also shows the resulting principal components from tangent PCA at the weighted Fr\'echet mean \cite{miolane2020geomstats} and from PNS. The method of tangent PCA first computes the mean value (the blue point) by minimizing the weighted distance from the mean to every data point. Then the method projects data onto the tangent space at the mean value. The succeeding PCA results in two principal components for the projected data. These two principal components are projected  onto the sphere, shown as the blue and the orange curve. This method fails to capture the variation of the data in a single mode. By contrast, PNS first fits a subsphere (i.e., $\mathbb{S}^1$) to the data so as to minimize the residuals. Then the method computes the Fr\'echet mean on $\mathbb{S}^1$ (shown as the red dot). Clearly, in this case, the backward mean from PNS is more representative of the population than the Fr\'echet mean (shown as the blue dot). Also, the fitted subsphere (shown as the black circle) efficiently represents the principal component of the data.

\begin{figure}[h!]
	\centering
	\includegraphics[width=\linewidth]{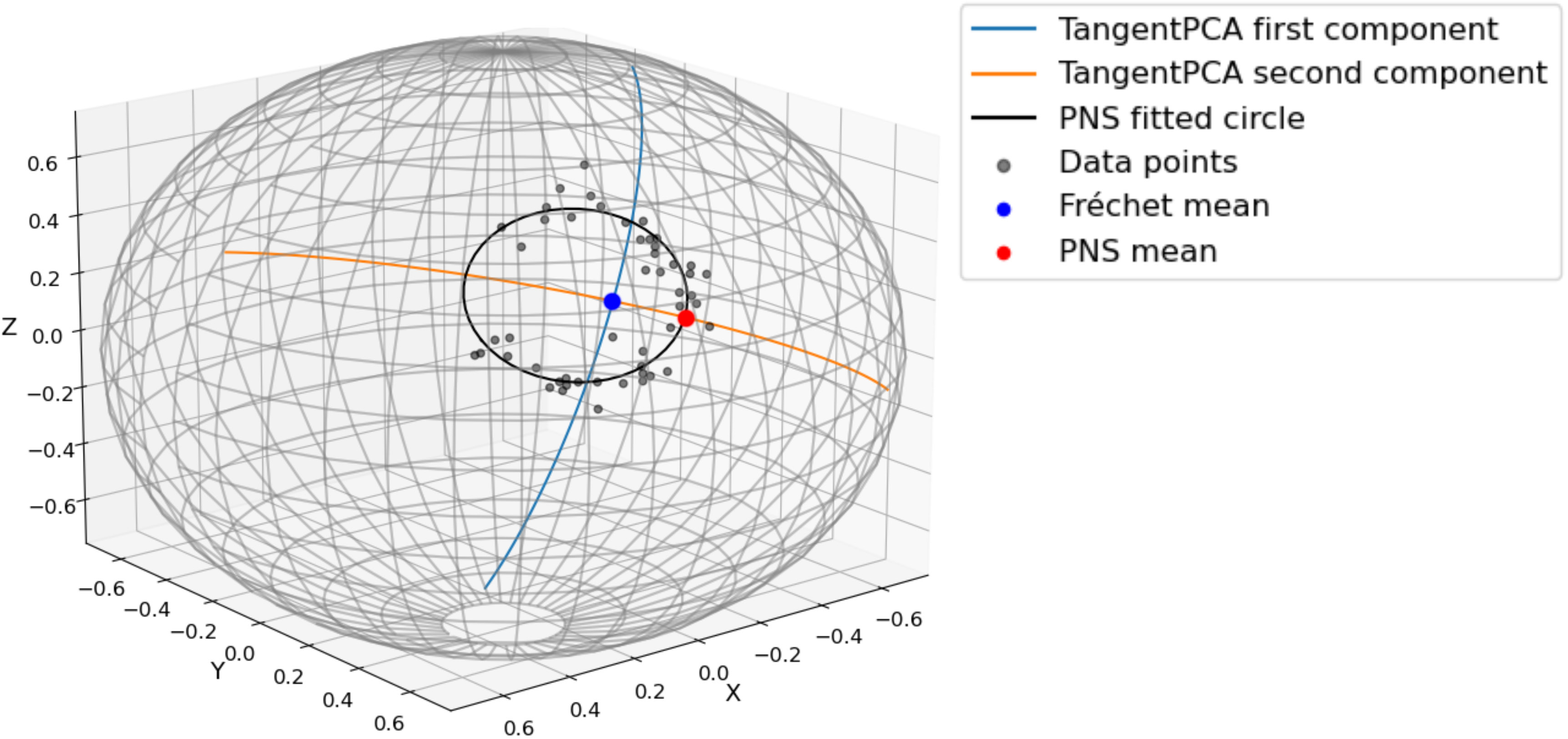}
	\caption{ Toy analysis of the non-Euclidean data (shown as the dark grey dots) simulated  via \cref{toy_exp_non_euclidean}. The principal components and mean values from two non-Euclidean methods. Principal Nested Spheres (PNS) results in more efficient representations of principal components (the black circle) and mean value (the red dot) than the method of tangent PCA at the weighted Fr\'echet mean (the blue dot) \cite{miolane2020geomstats}.}
	
	\label{fig_toy_non_euclidean}
\end{figure}
 
\end{example}

\subsection{Skeletal Representations}
\label{s_srep}
To facilitate the above statistical methods, a good geometric shape model is needed. A good shape model can (1) capture rich local shape information and (2) provide good correspondence for that shape information across samples. 
In representing 3D shapes, skeletal representations capture not only boundary geometry but interior shape information \cite{liyun2018,pouch2015medially}. Meanwhile, a generic skeletal representation consists of finitely many subsets of manifolds (also called strata \cite{damon2003smoothness}). Each of these strata exhibits consistent generic properties. Therefore the correspondence built upon these strata within a population can lead to desirable statistical results (see \cite{jp2016,schulz2016} for example). 

The Blum medial axis is one of the most well-known skeletal representations. A Blum medial axis for a 3D object is defined as a set of points that are centers of spheres which have one or more tangencies with the boundary. Siddiqi et al. \cite{siddiqi1999hamiltonian} proposed to construct the medial axis via the eikonal flow starting from the boundary. Although the resulting structures capture the boundary geometry of an object, a small perturbation on the boundary can lead to dramatic changes in the medial axis. Branches, for example, can be found due to small protrusions of the boundary. Such sensitivity can harm a succeeding statistical analysis.

Despite the sensitivity of a Blum medial axis, its spirit of representing shapes of regions has motivated research on radial geometry. For example, Damon \cite{damon2004} has found the relations between radial geometry and boundary differential geometry. Moreover, these relations can be approximately true in more general skeletal structures which satisfy a partial Blum condition, i.e., radial vectors are orthogonal to the boundary. 

An attractive property of general skeletal structures is the robustness against noise. S-reps, for example, generalized skeletal structures by not requiring Blum conditions to exactly hold \cite{steve2019,zhiyuan2020}. This relaxation of Blum conditions allows s-reps to have a consistent branching topology over the population. In addition to the consistent topology, s-reps have also shown several good statistical properties in anatomical shape analysis \cite{liyun2018}, including specificity, generalization and compactness. This means (1) (specificity) s-reps can generate objects that are similar to those in the training data (2) (generalization) s-reps have a good capability of representing unseen instances of the objects being studied and (3) (compactness) s-reps result in a tight distribution of a population of anatomical shapes. Due to these properties, s-reps have shown the superiority in classification \cite{jp_diss}, as compared to other landmark-based shape models (e.g., SPHARM-PDM \cite{Styner2006}).

	



\section{Methods}\label{sec_methods}
\subsection{Overview}
A statistical joint shape model for multi-object variation can be of importance for hypothesis testing and classification. To develop a good joint shape model, we combine the methods in shape analysis and a method in multi-block data analysis. Unlike other multi-object shape analysis methods, our method can simultaneously produce the joint and individual shape variation, thanks to the method AJIVE. In \cref{sec_jive} we detail this method of multi-block data analysis in the Euclidean setting. Further, PNS allows us to extend AJIVE to the non-Euclidean domain, yielding a novel method for multi-object shape analysis, as described in \cref{sec_pns}.

A challenge in multi-object shape analysis is alignment. The alignment of landmark-based shape models is promising. However, that alignment relies on the quality of correspondence across configurations. To obtain good correspondence across configurations for the alignment, we make use of s-reps and derive a shape model represented by landmarks from s-reps, as discussed in \cref{sec_sreps}. The resulting correspondence of the landmarks takes into account both the boundary and its interior geometry. 

\subsection{Joint Analysis of Multiple Blocks of Euclidean Data}
\label{sec_jive}
Our method to jointly analyze multi-block data decomposes the total variation into joint and individual variation. In this section, we focus on how the method works for multiple Euclidean data blocks \cite{feng2018}. Later, we discuss how to apply the method to multi-block data consisting of non-Euclidean data in \cref{sec_pns}. 

We assume in the decomposition that the joint variation is independent of the individual variation. It suffices to assume that the joint variation space is orthogonal to the individual variation space. 

Assume a block $X_k$ of dimension $d_k\times n$, where $d_k$ is the number of features of $X_k$ and $n$ denotes the number of configurations. Further assume each feature of $X_k$ is centered at zero. Then $X_k$ can be approximated by the total variation. We aim to decompose this total variation into (1) the joint component $J(X_k)$ whose variation patterns are shared by all blocks, (2) the individual component ${I_k}(X_k)$ that is specific to $X_k$ and (3) the additive residual $E_k$, i.e.,
\begin{equation}\label{eq_euc_ajive_eg}
    X_k = J(X_k) + {I_k}(X_k) + E_k, \qquad \forall k=1, \cdots, K
\end{equation}
where $J(\cdot)$ and ${I_k}(\cdot)$ denote, respectively, the projection of a block into the joint and the $k^{th}$ individual variation space. 

The key step to approach these projections is to estimate the basis vectors of the joint and the individual variation space. An important realization is that the joint and individual variation are each properties of scores. Hence, AJIVE decomposes the {score space} $ \mathbb{R}^n $ generated by $X_k$ into a joint variation subspace $J$ and its orthogonal complement called the individual variation space, i.e., 
 	\begin{equation}\label{eq_joint_space}
 	J \subseteq \mathbb{R}^n.
 	\end{equation}

 The subspace $J$, by definition, is shared by every block of data; namely, $J$ is the intersection of the score spaces of the blocks, assuming there is no noise. Specifically, we denote by $Q_k$ the score space of the $k^{th}$ block (we detail the construction of $Q_k$ later), assuming that $Q_k$ is of dimension $r_k$. Then, we have 
 \begin{equation} \label{eq_J}
     J\coloneqq\bigcap\limits_{k}Q_k \qquad\text{and}\qquad Q_k\subseteq \mathbb{R}^n.
 \end{equation}
 
It is unfortunate that due to noise in practice the joint variation space $J$ should be \textit{approximately} the intersection of $Q_k$'s. We assume that the space $J$ is of dimension $r$, where $r\leq r_k$. The computation of $J$ requires to find the $r$ basis vectors that are approximately shared by the $Q_k$'s. Those optimal basis vectors can therefore be found by minimizing the overall distance between the space $J$ and $Q_k$'s as follows. 
 
 Following \cite{feng2018}, we denote by $ \rho(\cdot, \cdot) $ the mapping from two Euclidean subspaces to their distance. The principal angles $\theta$'s of the two subspaces, e.g., $J$ and $Q_k$, are defined as acute angles $0 \leq \theta_1\leq \theta_2\leq \cdots\leq\theta_r\leq \frac{\pi}{2}$. These angles are evaluated for $i = 1, \cdots, r$ by
 \begin{equation}
     \theta_i =cos^{-1}\left(\max\limits_{v_j, q_k^j} \frac{<v_j, q_k^j>}{||v_j||\cdot||q_k^j||}\right) \qquad v_j \perp v_i, q_k^j \perp q_k^i
 \end{equation}
 where $<\cdot, \cdot>$ denotes the dot product and $\perp$ denotes orthogonality; $j=1, \cdots, i-1$. The vectors $v_j\in J$ and $q_k^j\in Q_k$ are called \textit{principal vectors} of the two subspaces. Principal Angle Analysis (PAA) yields the principal angles between two subspaces.
 The largest principal angle between the two subspaces is a measure of their distance \cite{stewart1990matrix,Miao1992}. For example, $\rho(J, Q_k)\coloneqq max(\{sin(\theta_i)\})$. 
 
The objective of AJIVE is to optimize the basis vectors of $J$ such that the overall distance between the space $J$ and the score spaces is minimized. We use $v_i$ to denote the basis vectors of $J$, where $i = 1, \cdots, r$. The minimization over the basis vectors can be written as 
  \begin{equation}\label{eq_dist}
     \argmin\limits_{v_i \in J}\sum\limits_{k}\rho(J, Q_k).
 \end{equation}
 In the following, we detail how AJIVE achieves a robust solution of \cref{eq_dist}.

\textbf{Setup.} The data in the joint analysis consists of multiple blocks $X_1, \cdots, X_K$ of Euclidean data. Each block $X_k$ is a matrix of dimension $d_k\times n$. Every block is organized in a way that the $i^{th}$ column of $ {X_k} $ represents the $k^{th}$ object in the $i^{th}$ configuration. AJIVE decomposes each block ${X_k}$ into the joint, individual and residual components (see \cref{eq_euc_ajive_eg}) such that the joint variation space satisfies the objective \cref{eq_dist}. 




\textbf{Low rank approximation.} The first step of the decomposition is to obtain approximately noise-free blocks using a low rank approximation of every block. In other words, $E_k$ in \cref{eq_euc_ajive_eg} should be identified and removed at this stage.

Let $ U_kS_kV^T_k $ be the Singular Value Decomposition (SVD) of the block $ X_k\in \mathbb{R}^{d_k\times n} $, where $ U_k$ and $ V_k^T$ are orthonormal matrices. To remove $E_k$, we assume that $X_k$ is of low rank $r_k$, where $ r_k\leq n $. Then the largest $r_k$ singular values in $ S_k $ are reserved in the low rank approximation. Also, we retain the corresponding $r_k$ columns (resp., rows) of the left (resp., right) singular matrix $ U_k $ (resp., $ V_k^T $). As a result, we obtain the approximated left singular matrix $\hat{U_k}\in \mathbb{R}^{d_k\times r_k}$, the approximated singular value matrix $\hat{S_k}\in \mathbb{R}^{r_k\times r_k}$ and the approximated right singular matrix $\hat{V_k}^T\in \mathbb{R}^{r_k\times n}$. The low rank (or noise-free) approximation of $X_k$ can be written as $\hat{X_k} = \hat{U_k}\hat{S_k}\hat{V_k}^T$. 

\textbf{Estimate joint variation subspace.} The rows of $ \hat{V_k}^T $ span the score space ${Q}_k$ of $\hat{X_k}$. The matrix  $\hat{V_k}^T$ is thus a representation of the score space ${Q}_k$ of the $k^{th}$ block. AJIVE calculates the principal angles based on SVD of the concatenation of $\hat{V_k}^T$'s. The resulting singular values are cosines of the principal angles, while the right singular vectors (also in $\mathbb{R}^n$) can be taken as the basis vectors of $J$ if the corresponding principal angles are small. The optimal space $J$ is thus spanned by the selected $r$ right singular vectors, where $r\leq r_k\leq n $. 

Both the dimension $r$ and the basis vectors of $J$ are the results that optimize \cref{eq_dist}. Essentially, the dimension depends on the degree of correlation between blocks. In particular, $ r=0 $ when there is a degenerate joint variation subspace, i.e., there is no significant joint variation between blocks.


\textbf{Individual variation.} Also, we can obtain the orthogonal complement of the space $J$ within each $Q_k$. This orthogonal space is called the individual variation space $I_k$. 

\textbf{Construction of the components.} Provided the basis vectors of the joint variation space $J$ and the individual variation space $I_k$, we are able to construct the joint structure $J(\hat{X_k})$ of $X_k$ and its orthogonal complement ${I_k}(\hat{X_k})$ given the low rank approximation $\hat{X_k}$ of $X_k$. 

We use $\hat{J}\in\mathbb{R}^{r\times n}$ to denote the basis matrix of the space $J$, i.e., each row of $\hat{J}$ is a basis vector. The projection of $X_k$ onto $J$ is written as the matrix multiplication $\hat{X_k}\cdot \hat{J}^T$. We note that this projection is indeed a transformation of the score space and thus is a \textit{right} multiplication. The $d_k\times n$ representation of the data $X_k$ projected in the joint variation space $J$ is $J(\hat{X_k}) = \hat{X_k}\cdot \hat{J}^T\cdot\hat{J}$.

We remark that the matrix $J(\hat{X_k})\in\mathbb{R}^{d_k\times n}$ is an even lower rank approximation of $X_k$. The rank of $J(\hat{X_k})$ is the rank of $\hat{J}^TJ$, i.e.,
\begin{equation}\label{eq_ranks}
rank(J(\hat{X_k})) = rank(\hat{J}^T\hat{J}\,)=rank(\hat{J}\,)=dim(J) =r
\end{equation}

We use this {joint structure} $ J(\hat{X_k}) $ in the classification model \cref{eq_classification}. The joint structure contains the integrated information of the configurations. We expect that the variation of the joint structure reflects the configuration-level variation due to common factors. Moreover, because this operator integrates information from multiple blocks, the variation of the joint structure is expected to be more robust against noise. In short, the operation $J(\cdot)$ should enhance the discriminatory power of individual blocks. 

Further, the individual component $I_k(\hat{X_k})$ is obtained via $I_k(\hat{X_k})=\hat{X_k}-J(\hat{X_k})$. The component $I_k(\hat{X_k})$ can also contain significant information of the multi-block data. We leave the analysis of these components as future research.

\subsection{NEUJIVE for Multi-object Shape Analysis}
\label{sec_pns}
 
In the context of multi-object shape analysis, each configuration is composed of the same number of objects, the joint shape variation describes how the neighboring objects vary together. The individual shape variation, on the other hand, describes how each object varies regardless of other objects in the configuration. Due to the manifold structure of shapes, we introduce a novel method in this section that simultaneously yields the joint and the individual shape variations in non-Euclidean contexts. 

Like the multi-block Euclidean data, each object in a multi-object shape dataset has a block of data associated with a score subspace. However, it is questionable to directly apply the above Euclidean AJIVE in multi-object shape analysis for the following reasons. (1) It is challenging to define principal components of shapes and the corresponding scores because of the manifold structure of shapes. (2) It is not straightforward to center shapes. (3) The linear operations that underlie \cref{eq_euc_ajive_eg} are not even defined on a curved manifold.


To address these problems, we propose to incorporate techniques in non-Euclidean statistics. In particular, we Euclideanize every object using their intrinsic PNS scores \cite{sungkyu2012}. It is helpful that PNS  yields a representative mean shape and efficient representations of principal components on the manifold.

PNS is suitable for the shape feature vectors that are in the pre-shape space with spherical geometry. As described in detail below, this method (1) fits a hierarchy of subspheres (treated as generalized principal components) to data by minimizing residuals and (2) uses the geodesic distances w.r.t. the fitted subspheres as the scores of the shapes. 

 We detail in the next section how we map the landmark-based shape models into the pre-shape space. Here, we assume that on each shape there are $ M $ boundary points in the 3D ambient space. Since each multi-object configuration contains $ K $ objects, we represent a multi-object configuration as a concatenation of $K$ feature vectors: $ \{x_i\in\mathbb{R}^{3\times M}\mid i=1, \cdots, K\} $. Each vector contains the boundary locations of a shape of an object. Altogether, multi-object configurations are represented by a matrix that stacks the $K$ blocks as described in \cref{sec_jive}.

As noted, the pre-shape space of single objects can be constructed by removing the location and size of shapes. Via Procrustes alignment, each shape can be transformed into a data point on the unit hypersphere and be centered at the origin \cite{dryden2016statistical}. This alignment also optimizes the orientation of a shape such that the shape is closest to the procrustes mean w.r.t. the metric in the pre-shape space.

\begin{figure*}[h!]
	\centering
	\boxed{\includegraphics[width=0.9\textwidth]{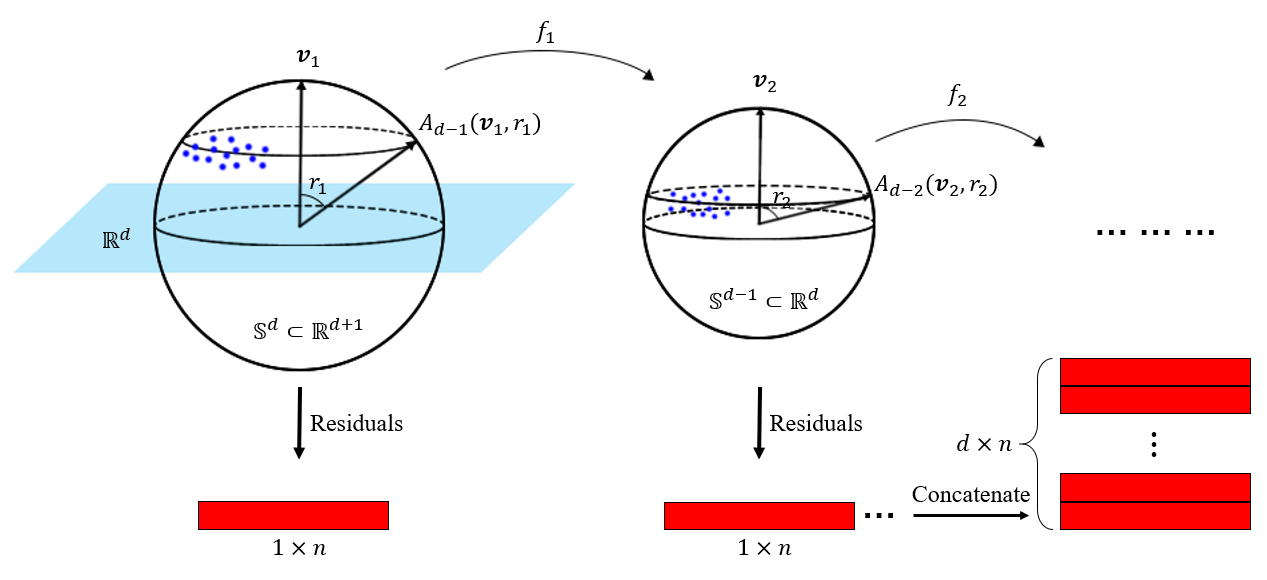}}
	\caption{Fitting PNS to $n$ samples in the pre-shape space of $d+1$ dimension. Blue dots illustrate pre-shapes on hyperspheres. Each subsphere $ A_{d-l} $ is parametrized by $ (v_l, r_l) $. The mappings $ f_i $'s are transformations from $ A_{d-i} $ to $ \mathbb{S}^{d-i} $.}
	\label{fig_pns}
\end{figure*}

\textbf{Fit subspheres to data.} Instead of fitting geodesics to single object shapes \cite{fletcher2004principal,huckemann2010intrinsic}, PNS fits a hierarchy of subspheres that are of decreasing dimensions from $ \mathbb{S}^{d-1} $ down to $ \mathbb{S}^0 $, as shown in \cref{fig_pns}. At each dimension, the best fitting subsphere is obtained by minimizing the sum of squared residual geodesic distances along the surface of the sphere. This best fitting subsphere is not necessarily a great circle such that the distribution along small circles can also be well represented \cite{damon2014backwards}. PNS computes the Fr\'echet mean on the nontrivial subsphere of the lowest dimension (i.e., $ \mathbb{S}^1 $). This Fr\'echet mean through the backward approach is typically more representative of the data (see \cref{fig_toy_non_euclidean}).

\textbf{Compute scores from the fitted subspaces.}
Since the pre-shape space has spherical geometry, we compute the signed geodesic distances from the data to the subspheres as the scores of the data. The corresponding PNS score space can typically be  treated as a Euclidean space.

As a result, each object is separately mapped to the Euclidean subspace $\mathbb{R}^{l_k}$, where $l_k$ is the number of the principal subspheres of the $k^{th}$ object. The resulting manifold for multi-object shape data can thus be expressed as the Cartesian product of the $K$ Euclidean subspaces, i.e., $\mathbb{R}^{l_1}\times \cdots\times\mathbb{R}^{l_K}$.


Our method extracts the joint shape variation from this Cartesian product of score subspaces of objects. 
 We achieve a good estimation of the joint shape variation  with the method discussed in \cref{sec_jive}. In essence, we have modified the decomposition model \cref{eq_euc_ajive_eg} to fit non-Euclidean data as follows.
\begin{proposition}\label{prop_pullback}
	Let $ X_k $ represent the $k^{th}$ object. The operator $ \phi_k $ represents the Euclideanization. The corresponding pullback operator $\phi_k^{\dagger}$ converts the Euclidean data in $\mathbb{R}^{l_K}$ to the landmark space $\mathbb{R}^{d_k}$, i.e.,
\begin{equation}\label{eq_inv_phi}
\phi_k^{\dagger}: \mathbb{R}^{l_k}\mapsto\mathbb{R}^{d_k}
\end{equation}
where $l_k < d_k$ in our context.
Then, we decompose $X_k$ with the composition of PNS, AJIVE and the pullback operator, i.e., 
	\begin{equation}\label{eq_generalized_decomposition}
	X_k = \phi_k^{\dagger}(J(\phi_k(X_k))+ I_k(\phi_k(X_k))+E_k(\phi_k(X_k)))
	\end{equation}
\end{proposition}

The pullback operation in \cref{eq_generalized_decomposition} allows us to interpret the discovered NEUJIVE modes of variation in terms of variations of landmarks, although we train the classification model \cref{eq_classification} on the Euclidean data $J(\phi_k(X_k))$. \Cref{algo_neujive} gives the procedure of computing the joint and individual components of $X_k$. The resulting joint structures that we concentrate on represent integrated neighboring shape information.

\begin{figure*}[t]
	\centering
	\includegraphics[width=.98\textwidth]{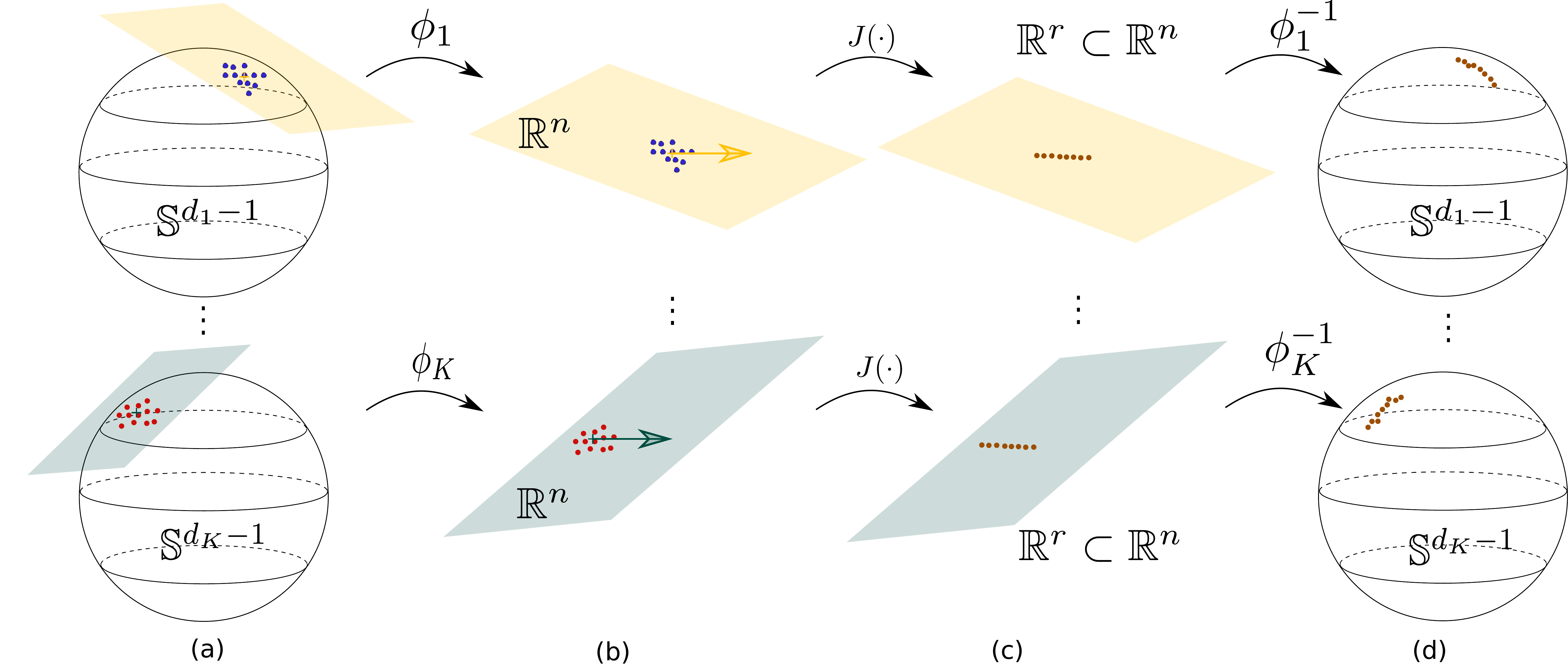}
	\caption{The intuitive illustration of extracting the joint components with NEUJIVE. (a) Input data live in non-Euclidean pre-shape spaces. (b) Construct the shared variation subspaces ($ \mathbb{R}^r $ shown as the yellow and green arrows) within the score spaces ($ \mathbb{R}^n $ shown as the yellow and green planes). (c) Project the data to obtain the joint components, shown as brown dots. Here, $ r $ denotes the dimension of the joint variation subspace. (d) Map the joint components (shown as the brown dots) of $X_k$ back to the pre-shape space via $\phi_k^{-1}$ for each $k$.}
	\label{fig_neujive}
\end{figure*}

\Cref{fig_neujive} illustrates the framework of NEUJIVE. We Euclideanize each object via $\phi_k$, as shown in \cref{fig_neujive} from column (a) to column (b). Then we estimate the joint variations from multiple Euclidean blocks. The estimated joint operator $J(\cdot)$ is applied to each separate Euclidean block, as shown in \cref{fig_neujive} from (b) to (c).
\Cref{fig_neujive} (d) illustrates the mapping of the joint structures to the pre-shape space.

\begin{algorithm}
  \caption{NEUJIVE for Multi-object Shape Analysis}\label{algo_neujive}
  \begin{algorithmic}[1]
    \REQUIRE Blocks $\{X_k|X_k\in\mathbb{S}^{(d_k-1)\times n},k=1, \cdots, K\}$
    \FOR{$k \leftarrow 1$ \algorithmicto{} $k$} \COMMENT{Euclideanization}
    \STATE $Z_k \leftarrow \phi_k(X)$ 
    \ENDFOR
    \STATE $Z\leftarrow[{Z_1}^T, \cdots, {Z_K}^T]^T$
    \STATE $Z_k \leftarrow J(Z_k) + I_k(Z_k) + E_k$ \COMMENT{AJIVE decomposition for each $k$}
    \RETURN $J(Z_k)$ and $I_k(Z_k)$
  \end{algorithmic}
\end{algorithm}

\subsection{Align Multi-object Landmarks from S-reps}
\label{sec_sreps}
This section discusses the shape model we use for multi-object shape analysis. Both landmark-based and skeletal-based models can capture boundary geometry. On one hand, landmarks directly sampled on boundaries allow us to align shapes according to the Euclidean distance between corresponding landmarks. On the other, skeletal-based models can capture richer shape information (e.g., interior geometry) with good correspondence, as explained in \cref{s_srep}. To provide good correspondence and to facilitate point-to-point alignment, we use skeletally implied landmarks in this research.

\begin{figure*}[h!]
	\centering
	\includegraphics[width=\linewidth]{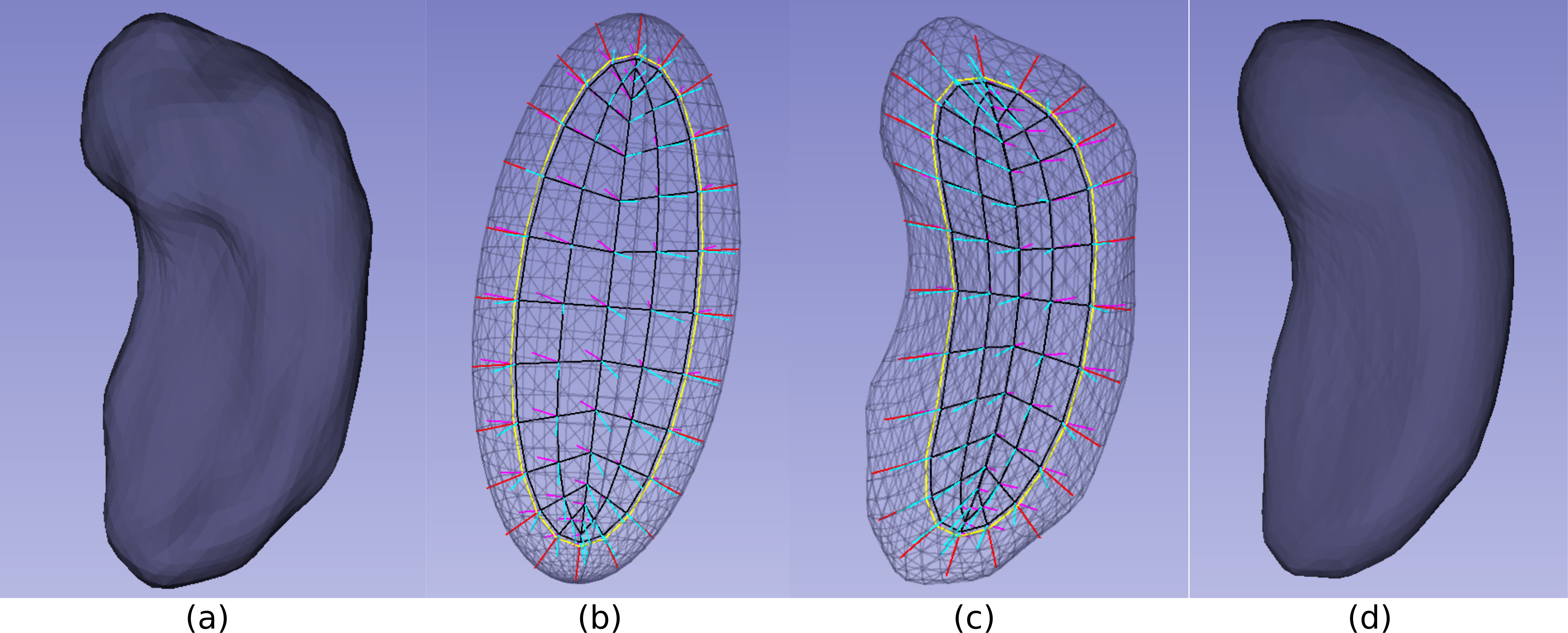}
	\caption{S-rep fitting to a hippocampus given a surface mesh that is shown in (a). The surface mesh is deformed into (b) a near ellipsoid (black transparent mesh) to which an s-rep (consists of a black grid and colorful spokes) is fitted. The correspondences of spokes are established via consistent discretization across cases. The s-rep is deformed back to  fit the hippocampus, as shown in (c). Figure (d) is the implied boundary of the fitted s-rep \cite{zhiyuan2020}. }
	\label{fig_ellipsoid_srep}
\end{figure*}

Considering the objects in this work have non-branching boundaries, we choose to fit non-branching s-reps \cite{zhiyuan2020} to single objects. The key idea of this fitting is to (1) deform a boundary of an object to an ellipsoid (2) compute and discretize the s-rep (also the medial axis) of the ellipsoid and (3) deform the discrete s-rep of the ellipsoid back to fit the object (see \cref{fig_ellipsoid_srep} (c)). 

Via interpolation, we are able to have a smooth s-rep. A smooth s-rep consists of a set of skeletal sheets and smooth radial vector fields (also known as \textit{spokes}) defined on each sheet. In particular, a (non-branching) skeleton contains two co-located skeletal sheets $M_+$ and $M_-$ and a fold curve $M_0$, which  bounds the two sheets. \Cref{fig_ellipsoid_srep} (b) shows the s-rep of an ellipsoid: $M_+$ and $M_-$ are shown as the black grids inside the ellipsoid, while the fold curve $M_0$ is shown as the yellow curve around the black grid.

The smooth field of spokes can map the skeletal sheets and the fold curve to an interior level surface and to form the implied boundary $\mathcal{B}$ of the object (see \cref{fig_ellipsoid_srep} (d)). The smoothness of spokes means a smooth vector field with no crossing spokes on each of the strata $M_+$, $M_-$ and $M_0$. This smoothness has been considered in the fitting process via the s-rep conditions (see \cite{damon2003smoothness,zhiyuan2020} for more details). Therefore, the fitted s-reps can produce smooth implied boundaries, via the smooth vector field of spokes
\begin{equation}
    S: \mathcal{M} \mapsto \mathcal{B}
\end{equation}
 where $\mathcal{M}$ denotes the disjoint union of $M_+$, $ M_-$ and $M_0$. Given a skeletal point $p\in \mathcal{M}$, we can obtain the corresponding implied boundary point via $S(p) \in \mathcal{B}$.

 Thus, the correspondence of the implied landmarks on $\mathcal{B}$ can be defined via the correspondence in sampling $S$, namely, the correspondence of spokes.
We sample spokes on each of $M_+$, $M_-$ and $M_0$ with a consistent pattern in the ellipsoids. An example is shown in \cref{fig_ellipsoid_srep} (b). Further, we establish the correspondence of spokes with the relative positions on the skeleton of the ellipsoids. Moreover, we maintain such correspondences during the deformation back to the s-rep of the object. The ends of the deformed spokes are taken as the corresponding implied landmarks of the object.

 The line segments in red, cyan and magenta in \cref{fig_ellipsoid_srep} are sampled spokes. The cyan spokes attached to $M_+$ point to a half of the boundary; the magenta spokes attached to $M_-$ point to another half of the boundary; the red spokes attached to $M_0$ point to the crest curve of the boundary for an ellipsoid.



The sampled spokes and the interpolation allow us to construct the implied landmarks of objects with good correspondence. We now consider the alignment of the implied landmarks in the pre-shape space. As discussed in \cref{sec_shape_space}, we can align single objects based on Euclidean distance between pairs of corresponding landmarks.  In multi-object configurations, however, each object should be separately considered in the alignment.

Based on the single object Procrustes alignment, we end with $K$ pre-shape hyperspheres. The shapes that lie on these hyperspheres are at the optimal position and orientation in terms of minimizing the Procrustes distance of an object. We apply \cref{algo_neujive} to these aligned shapes to extract the joint structures.

\section{Evaluation}
\label{sec_ev}
This section aims to validate the effectiveness and robustness of the proposed method with experiments on both simulated and real data. In \cref{s_toy}, we simulated two blocks of non-Euclidean data, each of which contains data that lie on $ \mathbb{S}^2 $. The simulated data are designed to be correlated across the two blocks. In \cref{sec_robust} we simulated two blocks of 2D landmarks shape data. Each block contains two groups. We show the benefit of the joint shape analysis in this toy classification problem. In section \cref{sec_real_data} we introduce the MRI brain images from the Infant Brain Imaging Study (IBIS) network. \Cref{sec_hypo_test,sec_exp_classification}, respectively, details the hypothesis testing and the classification of ASD and non-ASD on the brain structural shapes.
 Finally, we interpret the integrated shape information in the data space in \cref{sec_joint_interp}.

\subsection{Joint Structures of Manifold-valued Data}
\label{s_toy}
In this section, we aim to verify the effectiveness of our method for the joint analysis of multi-block non-Euclidean data. 

We simulate two correlated data blocks on $ \mathbb{S}^2 $. Each block contains $ 50 $ 3D points. They are generated to have a common joint structure parameterized by a uniformly distributed variable $ \theta\distas{} \text{Uniform}(0, \frac{3}{2}\pi)$. We generate some random noise for each block as independent multivariate Gaussian $ \epsilon_1\distas{i.i.d} N(0, 1)$ and $ \epsilon_2\distas{i.i.d} N(0, 1)$. The summations of the joint structure and the noise form a noisy circular distribution on a tangent plane centered at the north pole of the sphere; namely, $ a_ke^{i\theta} + \epsilon_k $, where $ k=1,2 $ representing the block index and $ a_k $ is a non-zero block-specific coefficient.
These data on the complex plane are then mapped onto $ \mathbb{S}^2 $ via an exponential mapping. Last, we rotate the data from the north pole to places that are specific to each block, which results in 
\begin{equation}\label{toy_exp}
X_k(\theta) = g_k(\varphi^{-1}(a_ke^{i\theta} + \epsilon_k))\qquad k=1, 2
\end{equation}
where $ \varphi^{-1} $ is an exponential mapping at the north pole; $ g_k $ is a block-specific rotation from the north pole on $ \mathbb{S}^2 $. 
\begin{figure*}[h!]
	\centering
	\includegraphics[width=\linewidth]{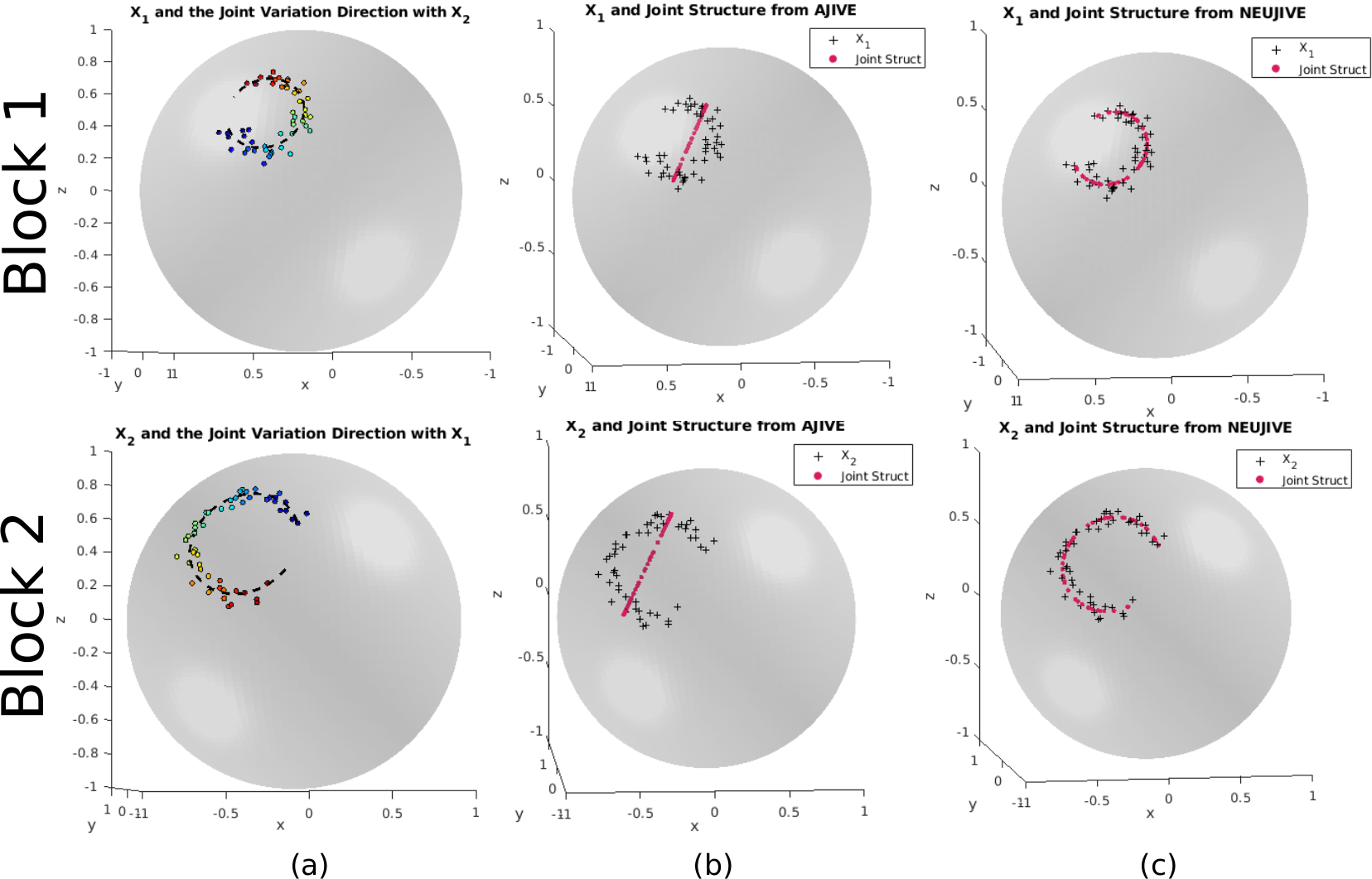}
	\caption{Column (a) shows simulated pairs of points on $ \mathbb{S}^2 $. The data tend to lie along a small circle with joint location on the circle indexed by the colors of points. Column (b) shows the joint structure (red points) estimated by AJIVE in $ \mathbb{R}^3 $. The AJIVE joint variation follows a linear pattern. Column (c) shows the joint structure estimated by NEUJIVE. This NEUJIVE joint structure gives a much more efficient representation of the joint variation pattern in the data.}
	\label{fig_toy_small_circle}
\end{figure*}

The column (a) in \cref{fig_toy_small_circle} shows the simulated two blocks $ X_1 $ and $ X_2 $ in the data space $ \mathbb{S}^2 $. $ X_1 $ and $ X_2 $ follow the respective small circles on $ \mathbb{S}^2 $ (as one moves along the rainbow colors). The two small circles are of different radii controlled by $ a_1 $ and $ a_2 $, respectively. Corresponding locations (i.e., $ \theta $'s) between the two circles are indexed by colors. 

As shown in \cref{fig_toy_small_circle} column (b), the joint structures estimated by Euclidean AJIVE are far from the underlying joint variation pattern. In particular, the red points lie along the first AJIVE direction following a linear pattern instead of a circular pattern.
In comparison, NEUJIVE can effectively extract more useful joint structures that follow a small circle pattern, i.e., provide a mode of variation that is much more descriptive of the actual variation in the data, as shown in  column (c) of \cref{fig_toy_small_circle}. 

This toy example demonstrates that our method can effectively capture the joint variation between multi-block non-Euclidean data. We owe this effectiveness to the faithful representation of scores of samples. Based on such faithful representations, NEUJIVE can result in more meaningful joint structures as compared to Euclidean AJIVE.
\subsection{Joint Shape Analysis in Classification}
\label{sec_robust}
In this section, we aim to verify the benefit from joint shape analysis in a classification problem. The key idea of this experiment is that it is sometimes difficult to classify two groups with close means. Yet, the  between-group variation is shared by multiple blocks of measurements. In this case, the joint variation can significantly improve the classification performance.

\begin{figure*}[h!]
	\centering
	\includegraphics[width=.98\linewidth]{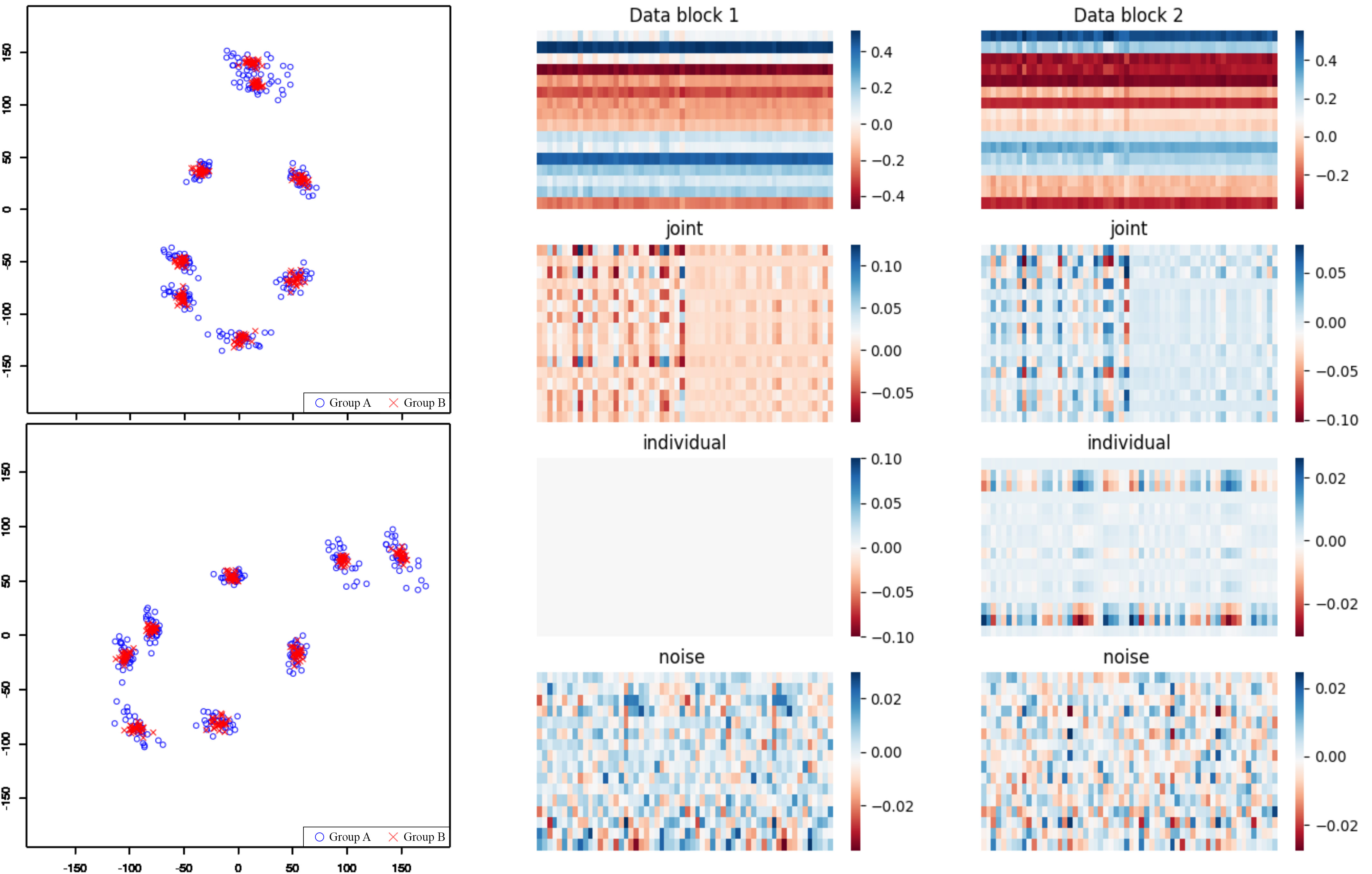}
	\caption{Joint analysis of two blocks (top left and bottom left figures) of 2D landmarks sampled from male gorillas' skulls. Each block has two groups (shown as the blue circles and the red crosses). We detail the generation of these landmarks in the text. The right two columns show the heat maps of the input matrices (the top row), the joint structures (the second row), the individual structures (the third row) and the residuals (the last row). The color represents the entry values of each matrix. Every matrix is organized in the same way, namely, rows are features while columns are samples (29 non-aligned samples and 29 aligned samples). The left half of every matrix contains features of non-aligned samples, while the right half contains features of aligned samples.}
	\label{fig_toy_classify}
\end{figure*}

\Cref{fig_toy_classify} shows an example of two data blocks, each of which has two groups. In this example, we start with a dataset from \cite{dryden2016statistical} in which eight 2D landmarks are sampled on each of the skulls of 29 male adult gorillas. The blue circles in the top left of \cref{fig_toy_classify} show the sampled landmarks. We treat the Procrustes aligned landmarks as another group, shown as the red pluses in the top left of \cref{fig_toy_classify}. These two groups have mean shapes that are very close. Thus, it is difficult to directly classify the two groups.  

Joint analysis via NEUJIVE extracts joint variation between multiple blocks. Hence, we created another block of landmarks that also contains two groups of landmarks formed by Procrustes alignment. First, we move the top-most landmarks in the top left of \cref{fig_toy_classify} to be farther away. In addition, we rotate all landmarks by 45 degree. These \textit{modified landmarks} are shown as the blue circles in the bottom left of \cref{fig_toy_classify}. Second, we obtain the Procrustes aligned landmarks of the modified landmarks. The aligned landmarks are shown as the red pluses  in the bottom left of \cref{fig_toy_classify}. 

The combination of the above two blocks of landmarks forms the input of NEUJIVE. Let $X_G\in\mathbb{R}^{16\times 29}$ represent the original landmarks of the gorillas' skulls. The  Procrustes alignment can be regarded as a function $\psi$ that results in a different group of data $\psi(X_G)$. The concatenation of these two groups is the block $X_1$ of the input of NEUJIVE. Moreover, we use $\widetilde{X}_G\in\mathbb{R}^{16\times 29}$ to denote the modified landmarks. Likewise, the  Procrustes alignment ${\psi}$ results in a different group of data ${\psi}(\widetilde{X}_G)$.  The two groups $\widetilde{X}_G$ and ${\psi}(\widetilde{X}_G)$ of the modified landmarks form another block $X_2$ of the input of NEUJIVE.
Therefore, the inputs of NEUJIVE are organized as 
\begin{equation}\label{eq_data_mat}
\begin{split}
X_1 = [X_G\quad \psi(X_G)]\in\mathbb{R}^{16\times 58}\\
X_2 = [\widetilde{X}_G\quad {\psi}(\widetilde{X}_G)]\in\mathbb{R}^{16\times 58}
\end{split}
\end{equation}
These two blocks have the same number of non-aligned and aligned cases. In the following, we first show classification with the joint structures of the two blocks from NEUJIVE. Then, we show classification of the non-aligned and the aligned group within each block, namely, $X_G$ vs. $\psi(X_G)$ and $\widetilde{X}_G$ vs. $\widetilde{\psi}(\widetilde{X}_G)$.

We obtain the pre-shapes of $X_1$ and $X_2$ by centering and normalizing each shape by the centroid size. The right two columns in \cref{fig_toy_classify} show the entry values of the matrices of the pre-shapes in the top row. The colors in the heat maps of the top row represent the entry values of the pre-shapes of $X_1$ (the middle) and $X_2$ (the right). The other rows show the entry values of the matrices resulting from the joint analysis of the pre-shapes of the landmarks with NEUJIVE. These matrices include (1) the joint components (shown in the second row), (2) the individual components (shown in the third row) and (3) the residual components (shown in the last row). 

 This  example is constructed in such a way that the two groups have different variations. This difference is hard to see in the input data matrices (the top row), the individual components (the third row) or the residual components (the last row). Yet, this group difference stands out very clearly in the joint components (the second row), in which the left half (from the non-aligned group) and the right half (from the aligned group) of the joint  components appear notable difference. It demonstrates that our NEUJIVE analysis of pre-shapes magnifies the shared between-group variation in the joint components. Therefore, it is clear that the improvement of classification follows from basing the analysis on the joint modes of variation. 

We utilize a robust linear classification method called Distance Weighted Discrimination (DWD) \cite{marron2007distance} for classifying the two groups, see the middle and the right column of \cref{table_toy_classify}. The classification of each block uses the following sets of features (see the rows of \cref{table_toy_classify}): (1) The original coordinates of landmarks (the first row). (2) The PNS scores (the second row). (3) The joint structures from AJIVE of the original coordinates (the third row). (4) The joint structures from AJIVE of the spherical coordinates (the fourth row). (5) The joint structures from NEUJIVE of the spherical coordinates (the fifth row). We repeatedly (100 repetitions) split the data for training and test. In each repetition, we randomly select 80\% of the data in the training while we use the remaining in the test. We report the average test accuracy over all the repetitions in \cref{table_toy_classify}. 

\begin{table}[h]\centering
\caption{Test performance using different features in classification}
		\label{table_toy_classify}       
\begin{tabular}{lcc}
\toprule
Features & {Male gorilla data block} & {Modified data block} \\
\midrule
Landmarks & 0.44  & 0.44\\
PNS & 0.5& 0.48  \\
Euclidean AJIVE & 0.59 & 0.55\\
Spherical AJIVE & 0.68  & {0.67}  \\
NEUJIVE & \textbf{0.75} & \textbf{0.72} \\
\bottomrule
\end{tabular}
\end{table}



Here, the Procrustes alignment in each block can be understood as an artificial driving force of the two groups. There exist biological factors (e.g., ASD) that change the distribution of shapes in a subtle way. In the following, we show the joint analysis of the structural shapes in human brains from an ASD dataset called the IBIS (short for Infant Brain Imaging Study) database.

\subsection{Autism Data for Multi-object Shape Analysis}
\label{sec_real_data}
We analyze MR brain images from 174 6-month-old infants. There are 33 of the children who were diagnosed as autistic later and 141 of these were shown not to have developed autism. From these images, the subcortical structures (i.e., the hippocampus, the caudate nucleus) were automatically segmented \cite{wang2014multi} and manually corrected by experts to produce label maps. 

We focus on configurations that consist of the left hippocampus and the left caudate nucleus. Given the label maps, we fit a triangular mesh to the boundary of each object using SPHARM-PDM \cite{Styner2006}. Though the vertices of these meshes provide correspondence in the spherical harmonic feature space, it is not straightforward to achieve geometric correspondence across cases. Thus, we fit s-reps to each object and extract 1002 implied boundary landmarks from the s-reps, as described in \cref{sec_sreps}. 

\subsection{Hypothesis Testing}
\label{sec_hypo_test}

The hypothesis testing aims to verify whether the difference between the two groups (ASD vs. non-ASD) is statistically significant in the joint variation subspace. Our inputs are $ K\times n $ implied PDMs, where $ K $ is the number of objects in each configuration and $ n $ is the number of samples. We propose to extract the joint components of these objects via \cref{algo_neujive} for the hypothesis testing, as follows. %

\begin{algorithm}[h!]
  \caption{Hypothesis testing with NEUJIVE}\label{algo_hypo}
  \begin{algorithmic}[1]
    \REQUIRE Blocks $\{(X_k, y_k) |X_k\in\mathbb{R}^{d_k\times n}, y_k=\{0, 1\},k=1, \cdots, K\}$
    \STATE $J \leftarrow NEUJIVE(X_k)$ \COMMENT{Apply \cref{algo_neujive}}
    \STATE $J^{y=1}, J^{y=0} \leftarrow J$\COMMENT{Partition data by class labels}
    \STATE $DiProPerm(J^{y=1}, J^{y=0})$ \COMMENT{Permutation test} 
    \RETURN \textit{pval} and \textit{zscore}
  \end{algorithmic}
\end{algorithm}

\begin{figure}[h!]\centering
	\includegraphics[width=\linewidth]{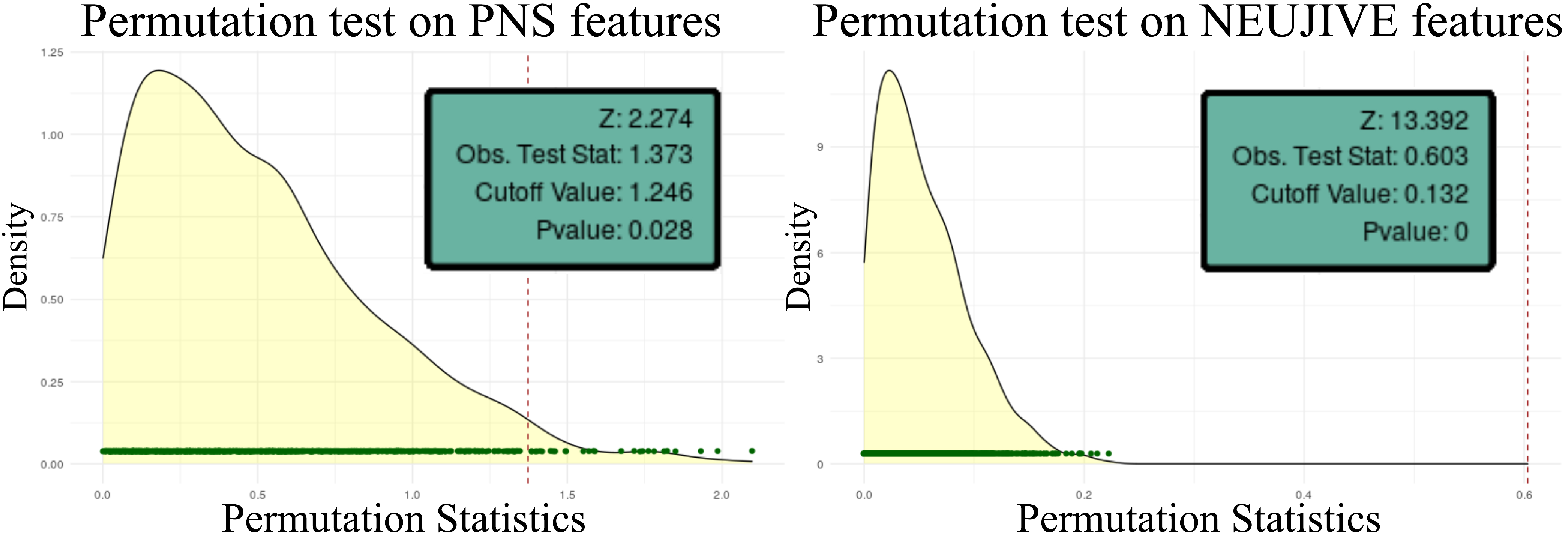}
	\caption{The results from the permutation test of PNS scores (left) and the test of NEUJIVE joint components (right). The green dots below each figure represent the test statistics under the permutations. The black curve represents the estimated density function of those group distances.}
	\label{fig_hypothesis_testing}
\end{figure}

In our experiments we found a joint rank of $ r=2 $ from \cref{fig_neujive} (c), which means that there are two significant joint variation directions. Within this 2-dimensional space, we test the data projected on the direction of the mean difference of the two groups with a permutation test, which is implemented by a method called Direction-Projection-Permutation (DiProPerm) \cite{wei2016direction}. In this setting, we simulate 1000 permutations and then compute the univariate statistic, i.e., mean difference (MD), between the two groups for each permutation.

\Cref{fig_hypothesis_testing} shows the test statistics on the MDs in those permutations using different sets of features. Moreover, the observed MD of the two groups in the joint variation space is at the positions of the brown dashed lines. Along with the permutation statistics, we show the empirical p-values, which are the proportion of the permutations that have bigger MDs than the observed MD, and the z-scores, which measure how many deviations that the observed MDs are above the average MD of their permutations. We use p-values and z-scores as the metric to evaluate the statistical significance of group difference in terms of the test features. Smaller p-values and bigger z-scores suggest more useful features for classifying ASD vs. non-ASD.

 \Cref{fig_hypothesis_testing} left shows the results from testing the PNS scores of hippocampal landmarks of the ASD and non-ASD configurations. There is a small proportion (2.8\%) of permutations that have larger MD between the ASD and non-ASD group than the observed MD. On the right of \cref{fig_hypothesis_testing}, we use the features from the NEUJIVE joint components in the test, as described in \cref{algo_hypo}. We found no permutations that have bigger MDs than the observed MD, i.e., the empirical p-value equals 0. Moreover, the z-score ($\approx 13.39$) is bigger than the z-score from PNS scores, meaning that the observed MD in terms of the NEUJIVE joint components is more statistically significant. The comparison between these two plots suggests that the joint structures can enhance the discriminatory power between the two groups. 

\Cref{table_hypo} presents more comprehensive results from the hypothesis testing using various sets of shape features. The first row shows the results from using the concatenation of s-reps implied boundary points. The second row shows the results from using the concatenation of Euclideanized PDMs with PNS. These two are commonly used ad-hoc methods for representing joint shape variations. The results suggest that  the Euclideanization improves the discriminatory power of the shape features. Moreover, using joint structures from AJIVE (the third row) results in a more statistically significant difference between the two groups. The fourth row shows the results of our proposed testing with NEUJIVE features. It demonstrates the advantages of using the joint shape variation in the testing. The bottom half of \cref{table_hypo} shows the results from the compared joint analysis methods that are variants of the existing methods reviewed in \cref{sec_back}. These methods can extract the correlation of features in high dimensional space. Yet, the results from these methods are not as strong as those from NEUJIVE. In particular, though the features selected by Hierarchical PLS (HPLS) also show statistically significant differences between the two groups, we found some DiProPerm anomalies that may occur due to overfitting, which we will study in later work. 
\begin{table}[h!]
	\centering
	\begin{threeparttable}%
		\caption{Hypothesis testing of ASD vs. non-ASD with different joint features}\label{table_hypo}
		\begin{tabular}{ccc}
			\toprule%
			& p-values  $\downarrow$\tnote{a}    & z-scores $\uparrow$\tnote{a}   \\
			\otoprule%
			Concatenate Implied PDMs     & 0.3106    & 0.3679           \\      
			Concatenate Euclideanized PDMs   & 0.028      & 2.274                 \\
			AJIVE \cite{feng2018}  & 0.0136     & 2.7714                 \\
			NEUJIVE (ours) & \textbf{0} & \textbf{13.392} \\
			\midrule
			HPLS \cite{Wold1996HierarchicalMP}   & 0.1135      & 1.206               \\
			PNS + HPLS & 0 & 4.4263  \\
			E-GCCA \cite{SHEN2014310} & 0.0169 & 2.0031 \\
			\bottomrule
		\end{tabular}
		\begin{tablenotes}
			\item[a] Smaller p-values and larger z-scores indicate more significant differences.
		\end{tablenotes}
	\end{threeparttable}
\end{table}
\subsection{Classification}
\label{sec_exp_classification}

In classification, the different dimensions between the training and test domains complicate the application of NEUJIVE. Let $ n_{tr} $ and $ n_{ts}$ be the number of training and test samples, respectively (oftentimes $ n_{tr}\neq n_{ts} $). Separate decomposition with NEUJIVE on the two domains will result in joint variation subspaces defined by vectors of different dimensions; i.e., $ J_{tr}\subseteq \mathbb{R}^{n_{tr}} $ while $ J_{ts}\subseteq \mathbb{R}^{n_{ts}} $. Thus, the learned classifier in $ J_{tr} $ cannot be applicable for the data in $ J_{ts} $.

To solve this problem, we project the concatenation of the training and test data using NEUJIVE into a joint variation subspace $ J\subseteq\mathbb{R}^{n_{tr} + n_{ts}} $. Then we train a classifier for the joint components of the training data, as described in \cref{algo_class}. 

\begin{algorithm}
  \caption{Classification with NEUJIVE}\label{algo_class}
  \begin{algorithmic}[1]
    \REQUIRE Blocks $\{(X_k, y_k) |X_k\in\mathbb{R}^{d_k\times n}, y_k=\{0, 1\},k=1, \cdots, K\}$
    \STATE $J \leftarrow NEUJIVE(X_k)$ \COMMENT{Obtain the joint components}
    \FOR{$h \leftarrow 1$ \algorithmicto{} $num\_holdouts$}  \COMMENT{Repeated hold-outs}
    \STATE $J_{tv}^{y=0}, J_{ts}^{y=0},  J_{tv}^{y=1}, J_{ts}^{y=1}\leftarrow J$\COMMENT{Random partition of the data}
    \STATE $J_{tr}^{y=0}, J_{val}^{y=0},  J_{tr}^{y=1}, J_{val}^{y=1}\leftarrow J_{tv}$\COMMENT{$J_{val}$ are used for cross-validation}
    \STATE Train the classifier (i.e., a linear DWD) on $J_{tr}^{y=0}, J_{tr}^{y=1}$ and cross-validate on $J_{val}^{y=0}, J_{val}^{y=1}$
    \STATE Test the classifier on $J_{ts}^{y=0}, J_{ts}^{y=1}$. Save the result ROC-AUC.
    \ENDFOR
    \RETURN Average of ROC-AUCs
  \end{algorithmic}
\end{algorithm}

The cross validation in \cref{algo_class} aims to estimate the initial rank $ r_k $ for the low rank approximation. With small $ r_k $'s (``under-ranks''), we can miss important geometric associations in the training data; with high $ r_k $'s (``over-ranks''), we may introduce too much noise to obtain an accurate decomposition. To achieve an optimal initial rank, we use the idea of \cite{SHEN2014310} called \textit{post-feature-selection} cross-validation to select $ r_k $ that optimizes the training performance.

Specifically, given the implied boundary points of the hippocampi and caudate nuclei, we first project all shapes into the joint variation subspace with NEUJIVE. Then we randomly partition the joint components of the hippocampi (or the caudate) from the ASD group into 10 roughly equal-sized
subsets and likewise with the non-ASD data. We set aside one of the subsets from each group for
testing and use the remaining subsets for training and validation. In this section we use the Area Under the ROC (Receiver Operating Characteristics) Curve (ROC-AUC) as the metric of performance.

\begin{figure*}[h!]\centering
	\includegraphics[width=\linewidth]{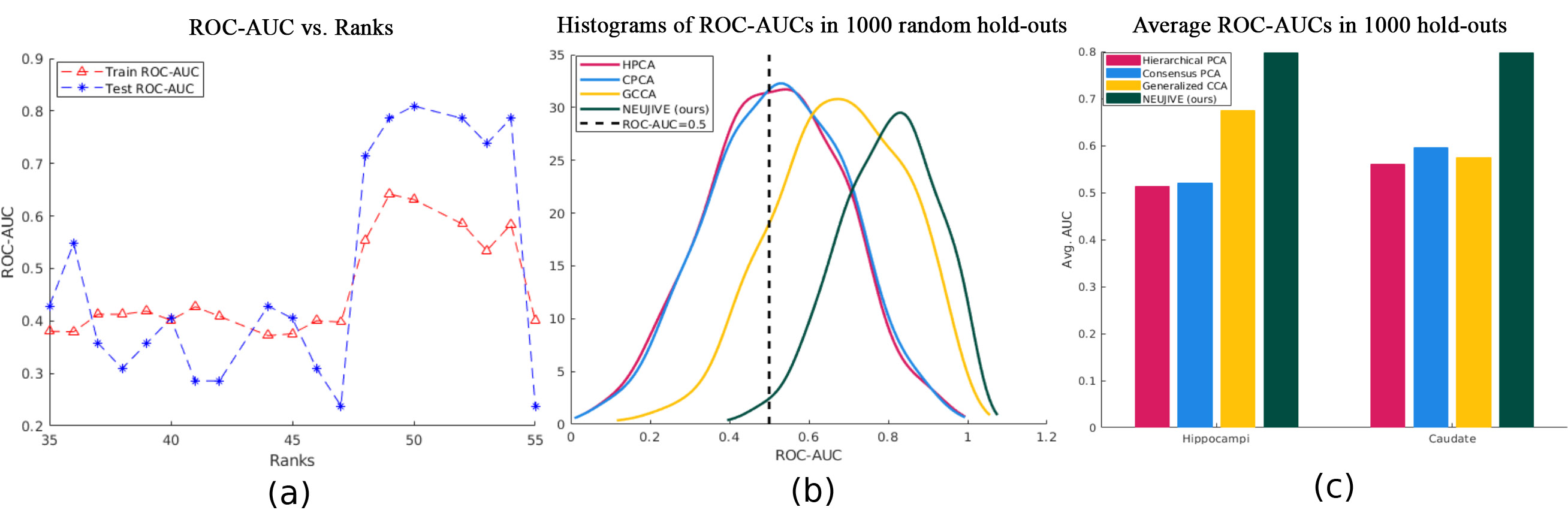}
	\caption{(a) The ROC-AUCs in classification of ASD and non-ASD with NEUJIVE joint features vs. the initial rank of the data block of the hippocampus. (b) Comparison of the histograms of ROC-AUCs from different methods. (c) The average ROC-AUCs of these compared methods using the joint components of hippocampi (left part) and caudate nuclei (right part).}
	\label{fig_auc_histogram}
\end{figure*}

\Cref{fig_auc_histogram} (a) shows the ROC-AUCs as a function of initial ranks for the block of the hippocampus data. From \cref{fig_auc_histogram} (a) we can see that the relation between the rank and ROC-AUC is not monotonic. However, the rank that maximizes the performance of training data can roughly maximize the test performance. \Cref{fig_auc_histogram} (a) also shows that both the under-ranks ($ r_k < 47 $) and over-ranks ($ r_k>54 $) can lead to poor performance in classification. 

To avoid bias in the testing procedure, we used the repeated random hold-out idea of \cite{jp_diss}. 
We conducted 1000 rounds of random hold-outs. We compared our method with other multi-block analysis methods, including Hierarchical PCA (HPCA) \cite{Westerhuis1998}, Consensus PCA (CPCA) \cite{Wold2005} and Generalized CCA (GCCA) \cite{SHEN2014310}. We show the histogram of the test ROC-AUCs in \cref{fig_auc_histogram} (b). In general, our method (shown as the green curve) performs better than the compared methods. 

\Cref{fig_auc_histogram} (c) shows the average test ROC-AUCs from our method (the green bars) and the compared methods. In classification using the joint structures of either the hippocampi or caudate, our method significantly improves the average test ROC-AUC. This improvement benefits from both the Euclideanization and the focus on joint structures provided by AJIVE.
\subsection{Shape Differences Between Groups in the Joint Variation Subspace}
\label{sec_joint_interp}
\begin{figure}[h!]\centering
	\includegraphics[width=.98\linewidth]{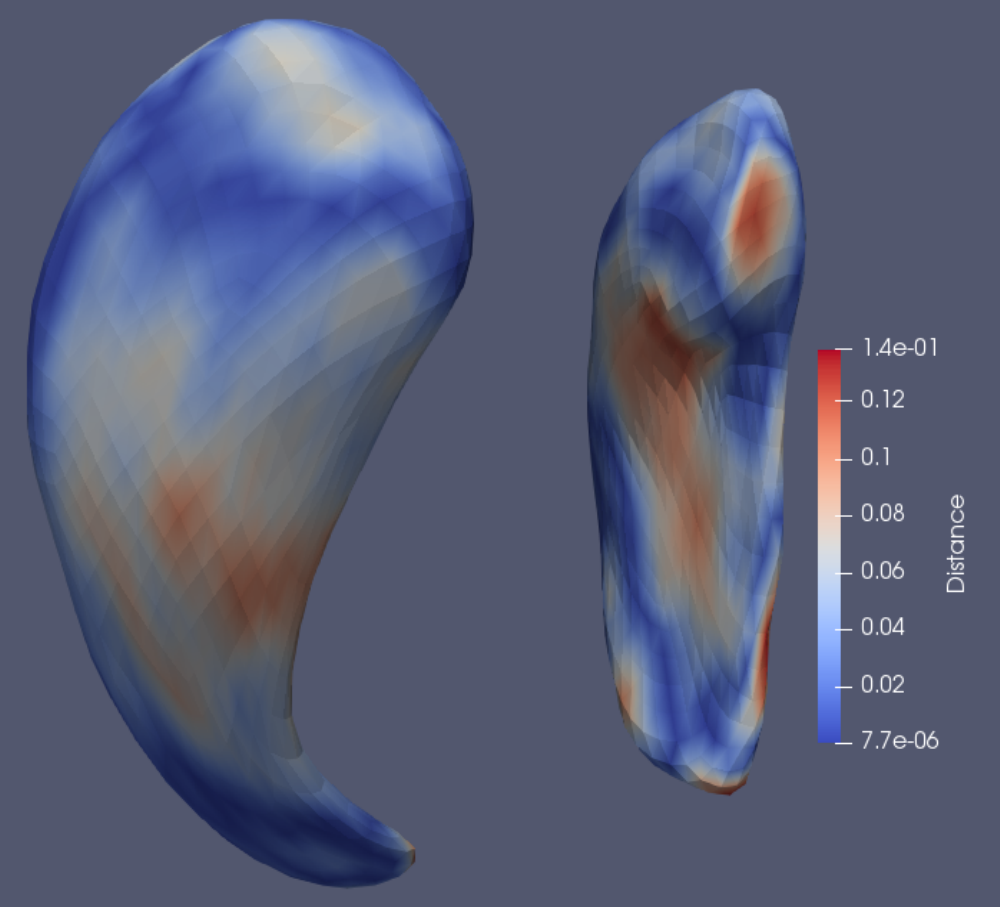}
	\caption{ The point-wise differences (in millimeters) between a reconstructed ASD configuration and a reconstructed non-ASD configuration. The reconstructed ASD and non-ASD configurations have more difference in the red regions than in the blue regions. }
	\label{fig_diff_dist}
\end{figure}

The learned classifier gives the boundary between the \textit{algorithmic} ASD (that were classified as ASD) and non-ASD (that were classified as non-ASD) group. To understand that group difference from the biological viewpoint, we highlight the localized algorithmic group differences on the corresponding subcortical structures. In particular, we interpret the algorithmic group difference relating to the joint variation direction. First, we average the scores of each of the two groups along the joint variation direction obtained from NEUJIVE. Second, we apply the pullback operations as described in \cref{prop_pullback} to the average joint scores w.r.t. each algorithmic group, resulting in landmarks for each object paired between the two groups. Three, we compute Euclidean distance between the landmarks of the algorithmic ASD group and those of the non-ASD group. 
The distances are shown as heat maps overlaid on a non-ASD configuration in \cref{fig_diff_dist}. 

Let the average joint score vector of the algorithmic ASD group be $\bar{J}_p$, where $p$ stands for positive. Likewise, the average score vector of the algorithmic non-ASD group is $\bar{J}_n$. We apply to the joint score vectors the pullback operations associated with the hippocampus $\phi^{\dagger}_H$ and with the caudate $\phi^{\dagger}_C$. This gives us good representatives of joint structures in the landmark space for each group. Each representative in the landmark space is a configuration that consists of a hippocampus and a caudate.

In \cref{fig_diff_dist} we show the heat maps of the Euclidean distances between the average positive configuration $(\phi^{\dagger}_H(\bar{J}_p), \phi^{\dagger}_C(\bar{J}_p))$ and the average negative configuration $(\phi^{\dagger}_H(\bar{J}_n), \phi^{\dagger}_C(\bar{J}_n))$. 
These heat maps suggest that some regions in the hippocampus and caudate are more different between the two groups than other regions. Some of these regions are consistent with the previous research on the relation between the development of ASD and regions of the subcortical structures (see e.g., the research on hippocampal regions of ASD data in \cite{dalton2019differences}).
 An interesting direction of future research would be to investigate the biological difference by incorporating covariates such as age and gender. 
\section{Conclusions and Discussion}
\label{sec_conclude}
This research aims to classify ASD and non-ASD with an integrative analysis of multiple brain structures. We hypothesized that joint shape analysis of the brain structures helps the classification based on the clinical findings that ASD can cause simultaneous shape changes of multiple brain structures. Thus, we developed a novel statistical method for multi-object shape analysis that yields additional insights beyond single shape analysis.

We noted that a desirable method should be suitable for medical applications in which a) sample sizes are small because each observation is expensive to acquire and b) a good interpretation is needed. These conditions require our method to be effective, robust and interpretable. With these requirements in mind, we extended AJIVE to a method for multi-block non-Euclidean data, called NEUJIVE. Our method can satisfy those requirements due to the following ideas. First, our method can faithfully represent multi-object shapes in disjoint Euclidean spaces with block-wise Euclideanization. This makes the following analysis more effective. Second, we studied space models and decided to focus on score spaces as compared to other integrative analysis methods (e.g., \cite{SHEN2014310,Wold1996HierarchicalMP}) which focus on feature spaces. This decision leads to a more robust method because i) score spaces are more stable in an HDLSS context and ii) such an integrative analysis method is robust against the differences among feature spaces. Third, our method of decomposition is non-parametric and is thus less likely to overfit the data. Last, we proposed to use s-reps implied boundary points that provide good correspondences across samples in the real data classification. 

We designed two toy examples studying multi-block non-Euclidean data analysis with NEUJIVE. For multi-block homogeneous data, NEUJIVE can effectively recognize the joint variation pattern (see \cref{s_toy}). For multi-block heterogeneous data (with group labels), NEUJIVE focuses on the group difference in the joint structures and thus results in higher classification accuracy (see \cref{sec_robust}). These toy examples explained the advantages of NEUJIVE on multi-block non-Euclidean data analysis.

We then carried out experiments, including hypothesis testing and classification, on the ASD data. The results from the hypothesis testing show that NEUJIVE features are more useful in distinguishing the ASD and non-ASD group than other joint features. Likewise, NEUJIVE also outperforms other joint feature learning methods in the classification. 

Although we focus on the joint shape variation in this research, the individual shape variation can also be useful for particular data and applications. We will explore the individual shape variation in future research.

Generally, the proposed method can be applied to non-Euclidean data that live on a polysphere (e.g., see \cite{Eltzner2015}) or the cross-products of manifolds (e.g., see \cite{steve2019}). To adapt to the different geometry of data domains, differing Euclideanization methods should be used.

This work has a few possible future directions. First, it is worthy of utilizing shape features inside objects that are provided by s-reps. Second, the separate alignment and Euclideanization in this work ignore the spatial interrelation between neighboring shapes. In the future, we will address this issue with improved multi-object shape models where the alignment and Euclideanization can characterize that interrelation. Third, we will apply the method to more datasets and with more structures involved. In addition, it would be of interest to research on the individual shape variation.


%

\begin{acknowledgements}
This research is funded by NIH grants R01HD055741, R01HD059854 and R01HD088125. The data for the applications was kindly provided by the IBIS network. We thank G. Gerig (NYU), SunHyung Kim (UNC), D. Louis Collins (McGill University), Vladimir Fonov (McGill University) and Heather Hazlett (UNC) for their help in processing the data. We are also grateful for the useful discussion relating to this project with Xi Yang, Iain Carmichael and Eric Lock.
\end{acknowledgements}

%
%

\bibliographystyle{spmpsci}      
\bibliography{neujive}   


\end{document}